%% file: root.tex
\documentclass[letterpaper, 10 pt, conference]{ieeeconf}  

\IEEEoverridecommandlockouts                              

\usepackage{graphicx}
\usepackage{amssymb}
\usepackage{amsmath}
\usepackage{bm}
\usepackage{xspace}
\usepackage{makecell}
\usepackage{booktabs}
\usepackage{url}
\usepackage{subfigure}
\usepackage{tabularx}
\usepackage{multirow}
\usepackage{multicol}
\usepackage{threeparttable}
\usepackage{placeins}
\usepackage{balance}

\usepackage[
    style=ieee,
    natbib=true,
    citestyle=numeric-comp,
    doi=false,
    isbn=false,
    url=false]{biblatex} 

\makeatletter
\let\NAT@parse\undefined
\makeatother
\usepackage{hyperref}

\newcommand{\ie}{i.e.,\ }
\newcommand{\eg}{e.g.,\ }

\newcommand{\etal}{\xspace{}et al.\xspace}

\usepackage{color}
\usepackage{xcolor}

\title{\LARGE \bf
Sim-to-Real Transfer for Quadrupedal Locomotion \\via Terrain Transformer
}

\author{Hang Lai$^{1,2}$, Weinan Zhang$^{1}$, Xialin He$^{1}$, Chen Yu$^{2,3}$ , Zheng Tian$^{4}$, Yong Yu$^{1}$, and Jun Wang$^{2,5}$
\thanks{$^{1}$Dept. of Computer Sci. and Eng., Shanghai Jiao Tong University, China.}
\thanks{$^{2}$Digital Brain Lab, Shanghai, China}%
\thanks{$^{3}$School of Info. Sci. and Tech., ShanghaiTech University, China.}
\thanks{$^{4}$School of Creativity and Art, ShanghaiTech University, China.}%
\thanks{$^{5}$Centre for Artificial Intelligence, University College London, UK.}
}

\begin{document}
	
\maketitle
\thispagestyle{empty}
\pagestyle{empty}

\begin{abstract}
Deep reinforcement learning has recently emerged as an appealing alternative for legged locomotion over multiple terrains by training a policy in physical simulation and then transferring it to the real world (\ie sim-to-real transfer). Despite considerable progress, the capacity and scalability of traditional neural networks are still limited, which may hinder their applications in more complex environments. In contrast, the Transformer architecture has shown its superiority in a wide range of large-scale sequence modeling tasks, including natural language processing and decision-making problems. In this paper, we propose Terrain Transformer (TERT), a high-capacity Transformer model for quadrupedal locomotion control on various terrains. Furthermore, to better leverage Transformer in sim-to-real scenarios, we present a novel two-stage training framework consisting of an offline pretraining stage and an online correction stage, which can naturally integrate Transformer with privileged training. Extensive experiments in simulation demonstrate that TERT outperforms state-of-the-art baselines on different terrains in terms of return, energy consumption and control smoothness. In further real-world validation, TERT successfully traverses nine challenging terrains, including sand pit and stair down, which can not be accomplished by strong baselines. 
\renewcommand{\thefootnote}{}

\end{abstract}

\input{introduction.tex}

\input{related_work.tex}
\input{preliminaries.tex}
\input{method.tex}
\input{evaluation.tex}
\input{conclusion.tex}
\input{ack.tex}
\printbibliography
\input{appendix.tex}
\end{document}

%% file: introduction.tex
\section{Introduction}
\label{sec:Introduction}
Legged locomotion over varied terrains is challenging due to the potential changing dynamics and irregular profiles, surfaces, and obstructions \cite{terrain}. To tackle this problem, deep reinforcement learning (RL) methods \cite{tan2018sim, terrain, rma, hwangbo2019learning, imitate} have been developed by directly training a controller in simulation with diverse terrains and environmental parameters and then transferring it to the real world using different heuristic adaptation techniques \cite{imitate, yu2019sim, yu2020learning, nagabandi2018learning, song2020rapidly}. Typically, in the well-known \emph{privileged learning} framework \cite{terrain}, a teacher policy will first encode the privileged information, \eg heightmaps and physical parameters, which are inaccessible directly in the real world, into a low-dimensional latent vector. Subsequently, a student policy is trained to recover such a latent vector by just using the previous proprioception sequences with conventional sequence models like temporal convolutional network (TCN) \cite{TCN, rma, terrain} or recurrent neural network (RNN) \cite{GRU, wild}.

Though privileged learning and its extension have been widely exploited and achieved remarkable success in quadrupedal locomotion over multiple terrains, it still has some limitations. Firstly, a versatile teacher policy will make full use of the latent vector, denoted as $l_t$, to represent the environmental property, making it difficult to be estimated online. To address this issue, Peng \etal~\cite{imitate} proposed to constrain the mutual information between $l_t$ and the privileged information, which, however, may impede the adaptability of the policy. Secondly, the learned student policy tends to be sensitive to the estimated $l_t$, and may suffer catastrophic degradation in performance if $l_t$ is out of its training distribution. Therefore, can we avoid the strong dependence on the precision of the estimated vector? 

\begin{figure}[tb]
	\centering
         \vspace{5pt}
	\includegraphics[width=0.49\textwidth]{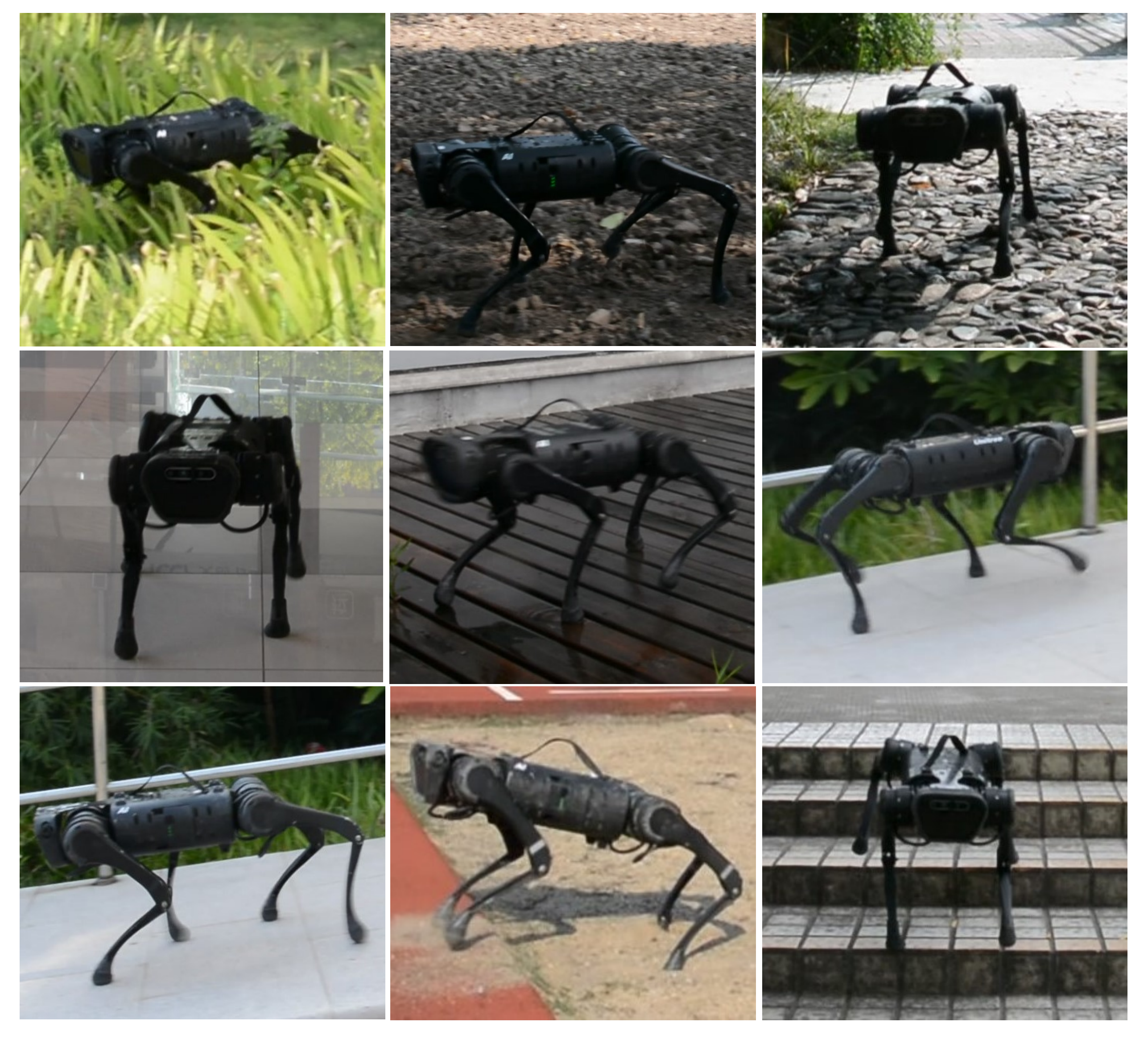}
	\vspace{-20pt}
	\caption{Application of TERT on A1 robot over multiple challenging terrains.}
	
	\vspace{-15pt}
	\label{fig:real world}
\end{figure}


One promising solution is to circumvent the encoded $l_t$ and directly predict the actions of the teacher policy conditioned on the previous proprioception sequences. We argue that this sequence prediction paradigm is exceptionally suitable for the Transformer architecture since it has demonstrated remarkable performance improvement and generalization on a wide range of sequence modeling problems due to its high capacity for modeling and robustness for decoding. In addition, the self-attention mechanism of Transformer could make better use of historical information through credit assignment to capture the characteristics and changes of the environment. Moreover, Transformer has proven to be able to model diverse behaviors \cite{ramesh2021zero}, which is crucial for transferring to unseen scenarios. Therefore, it is tempting to utilize the modern Transformer models to imitate the teacher policy's behavior over various terrains using the past trajectories, which can seamlessly take advantage of innovations in sequence modeling literature.

To this end, we propose Terrain Transformer (TERT), a novel Transformer model with privileged learning, and present a two-stage training framework explicitly designed for sim-to-real transfer. More specifically, in the \emph{offline pretraining} stage, the teacher policy interacts with a simulator and collects trajectories for TERT training. This stage resembles the standard Transformer training process in language and offline RL areas \cite{Transformer, DT, TT}. After the first stage, Transformer can give relatively accurate predictions conditioned on the teacher's observation-action sequences. However, the input sequence distribution of the Transformer will shift when deploying. Therefore, in the \emph{online correction} stage, TERT interacts with the simulator while the teacher simultaneously gives actions as the target. Then TERT is trained to fit the teacher's actions conditioned on its own observation-action sequences.

We compare TERT to the state-of-the-art baselines on different terrains both in simulation and in the real world. From the simulation experiments, TERT surpasses the baselines with higher return and less energy consumption. We then compare TERT and the standard student policy on a real Unitree A1 robot. In the real-world evaluation, TERT successfully traverses nine different terrains, including grass, soil, pebble, slope, sand pit, stair down, etc., as Figure~\ref{fig:real world} shows. While the student policy accomplished tests on other terrains, it failed on sand pit and stair down, verifying the advantage of Transformer architecture compared to conventional models on challenging terrains. To the best of our knowledge, this is the first work that deals with sim-to-real transfer of quadrupedal locomotion over multiple challenging terrains via Transformer-based sequence modeling, which could become a new paradigm for sim-to-real transfer tasks.



%% file: related_work.tex
\section{Related Work}
\label{sec:Related_Work}
\noindent \textbf{Sim-to-Real RL for Legged Locomotion.} Sim-to-Real reinforcement learning (RL) for legged locomotion has recently gained immense interest due to its potential to eliminate the need for human expertise in controller design \cite{tan2018sim, terrain, rma, imitate, hwangbo2019learning,da2020learning,yang2022fast, shi2022reinforcement}. However, it is non-trivial to transfer the policy trained in simulation to real robots due to the discrepancy between simulation and the real world, known as the \emph{reality gap} \cite{realitygap1, realitygap2}. Much effort has been devoted to narrowing this reality gap, either by increasing the simulator's accuracy \cite{tan2018sim, hwangbo2019learning} or by randomizing the physical parameters such as friction and mass during training (domain randomization) \cite{randomization1, randomization2, s2rsurvey, chebotar2019closing}. Other works are dedicated to introducing more inductive bias to facilitate the training and transferring of policy. For example, Peng \etal~\cite{imitate} trained a robot by imitating the poses of natural animals. Kumar \etal~\cite{rma}, instead, utilized bioenergetics to help design reward function. Our method is orthogonal to them and can naturally take advantage of the above techniques. Another related line of research is system identification, which tries to estimate the raw environmental parameters \cite{osi} or the encoded ones \cite{rma, terrain} from historical observations for online adaptation. 
\vspace{0.1cm}
\newline
\noindent \textbf{Transformers.} Recent advances in Transformers \cite{Transformer} have led to significant breakthroughs in a number of fields. In natural language processing (NLP) \cite{NLP1, bert}, Transformer has shown great superiority against the traditional LSTM \cite{lstm} or GRU \cite{GRU} models. Applications of Transformer have increasingly expanded to other areas, such as computer vision (CV) \cite{CV1,CV2, CV3} and RL. For example, in the RL area, Decision Transformer \cite{DT} utilizes an autoregressive model to generate action sequences conditioned on desired return and past state-action pairs. Trajectory Transformer \cite{TT}, instead, leverages Transformer to generate entire trajectories by discretizing states, actions, and rewards into tokens. Besides, Wen \etal~\cite{wen2022multi} proposed to model multi-agent RL as a sequence modeling problem and resorted to Transformer to solve it. Furthermore, the Gato model \cite{reed2022generalist} successfully trains and deploys one Transformer model on hundreds of tasks, including robot arm manipulation. Our method draws inspiration from Decision Transformer due to its simplicity and computational efficiency, which are critical in real-world applications.

One closely-related work is Yang \etal~\cite{yang2021learning}, which built a Transformer model to fuse information for quadrupedal locomotion. TERT differs from theirs mainly in the following three aspects: i) Different modeling - 
their Transformer takes the different kinds of information in a \emph{single timestep} as input and models the joint patterns at this moment. In contrast, TERT models quadrupedal locomotion as a sequence prediction problem and takes observations of \emph{multiple timesteps} in the history as input, thus better leveraging the superiority of Transformer in sequence modeling. 
ii) Different training paradigm - Yang \etal \cite{yang2021learning} trained the Transformer purely via online RL, while TERT adopts a novel two-stage training framework, which can naturally inherit the merit of privileged learning. iii) Different terrain difficulty - compared with Yang \etal~\cite{yang2021learning}, TERT exhibits a more agile and mild behavior and successfully traverses more challenging terrains such as sand pit and stair down, even without visual inputs, which are not included in their work.

%% file: preliminaries.tex
\section{Preliminaries}
\label{sec:Preliminaries}

\subsection{Reinforcement Learning for Quadrupedal Locomotion}
We formulate quadrupedal locomotion as a Partially Observable Markov Decision Process (POMDP) defined by the tuple ($\mathcal{S}, \mathcal{O}, \mathcal{A}, T, r, \gamma$), where $\mathcal{S}$, $\mathcal{O}$, and $\mathcal{A}$ are the state, observation, and action spaces, respectively. $T(s_{t+1}\mid s_t,a_t)$ is the transition density of state $s_t$ given action $a_t$, and the reward function is denoted as $r(s_t,a_t)$. $\gamma \in (0,1)$ is a discount factor. At each timestep $t$, only the partial observation $o_t \in \mathcal{O}$ can be observed instead of $s_t$ due to the limitation of sensors.  The goal of reinforcement learning (RL) is to find the optimal policy $\pi^* \colon \mathcal{O} \to \mathcal{A}$ that maximizes the expected return (sum of discounted rewards):
\begin{equation}
\label{eq: rl-obj}
\pi^* \!\colon\!\!\!\!= \mathop{\arg \max}_\pi \mathbb{E}_{s_{t+1} \sim T(\cdot \mid s_t,a_t), a_t \sim \pi(\cdot \mid o_t)} \Big[\sum_{t=0}^\infty \gamma^t r(s_t,a_t) \Big].
\end{equation}
\subsection{The Transformer Model}
Our model draws upon the Decision Transformer \cite{DT}. The difference is that we remove the returns-to-go in the trajectory representation since we find that including the returns-to-go will slightly degrade the performance of our algorithm in preliminary experiments. To be more specific, we use the GPT \cite{gpt} architecture consisting of $n$ stacked self-attention layers with causal masking. Give a trajectory: $\tau = (o_1, a_1, o_2, a_2, \dots o_T, a_T)$ of length $T$, The Transformer first embeds the inputs into $\left\{x_i\right\}_{i=1}^{2T}$ with position embedding. The self-attention layers then map each $x_i$ into the corresponding key, query and value, denoted as $k_i$, $q_i$, and $v_i$, respectively, through linear transformation. The $i$-th output of the self-attention layer is calculated as:
\begin{equation}
    z_i=\sum_{j=1}^{i} \operatorname{softmax}(\left\{\left\langle q_i, k_{j^{\prime}}\right\rangle\right\}_{j^{\prime}=1}^{i})_j \cdot v_j,
\end{equation}
where $\left\langle \cdot \right\rangle$ denotes the dot product operation, and the softmax normalization of the dot product represents the attention weight assigned to the $j$-th token. Note that only the tokens $j \leq i$ are used when calculating $z_i$ since we should ensure that the Transformer can only access preceding tokens to predict the next ones. The output of the last self-attention layer is then fed into another linear layer to predict the actions conditioned on the previous observation-action sequence:
\begin{equation}
    \hat{a}_t \sim \mathrm{Trans}(\cdot \mid o_1,{a}_1,\dots, o_t), \quad \forall t \in [1,T].
\end{equation}
The GPT Transformer is trained to minimize the mean square error between the predicted actions $\hat{a}_t$ and the target actions.
For evaluation, the Transformer takes the last $T$ observation-action pairs as input and executes the predicted $\hat{a}_T$ in the environment.

%% file: method.tex
\section{Terrain Transformer}
\label{sec:Method}


In this section, we introduce Terrain Transformer (TERT), a novel approach that combines high-capacity Transformer models with privileged learning through a two-stage training framework.
An overview of the training framework of TERT is illustrated in Figure~\ref{fig:framework}.


\subsection{Teacher Policy Training}
Like the standard privileged learning framework \cite{rma, terrain}, our method will first train a teacher policy in simulation with access to the privileged information, which mainly consists of three parts: i) elevation map around the robot base; ii) contact force with the ground; iii) ground-truth physical environmental parameters used for domain randomization, such as friction and mass. Following Kumar \etal~\cite{rma}, let $e_t$ denote the privileged information, which will first be processed into a latent vector $l_t$ by a multi-layer perceptron (MLP) encoder $\mu$. Then, the teacher policy $\bar{\pi}$ gives actions conditioned on $l_t$ and proprioception observation $o_t$: 
\begin{align}
    l_t &= \mu({e_t}),
    \\
    \bar{a}_t &= \bar{\pi}(o_t, l_t).
\end{align}
The teacher policy and encoder are trained jointly via PPO (Proximal Policy Optimization) algorithm \cite{ppo}.

\subsection{Terrain Transformer Training}
\label{sec:Transformer Learning}
\noindent \textbf{Offline Pretraining.} Although teacher policy can achieve near-optimal performance and successfully traverse different terrains in simulation, it can not be directly deployed in the real world since the privileged information is noisy or difficult to obtain \cite{wild}. Unlike previous methods that reserve the teacher policy and try to estimate $l_t$ online \cite{terrain, rma}, our method chooses to train a Transformer controller from scratch. Furthermore, we propose a two-stage training framework to better integrate Transformer with privileged learning. Precisely, in the first offline pretraining stage, the teacher policy is executed in simulation 
over varied terrains with privileged information to collect data (Figure~\ref{fig:framework} top-left). Then a GPT Transformer is trained on the collected data (observation-action sequences without privileged information) to predict the teacher's action (Figure~\ref{fig:framework} top-right):
\begin{figure}[t]
	\centering
 \vspace{5pt}
	\includegraphics[width=0.49\textwidth]{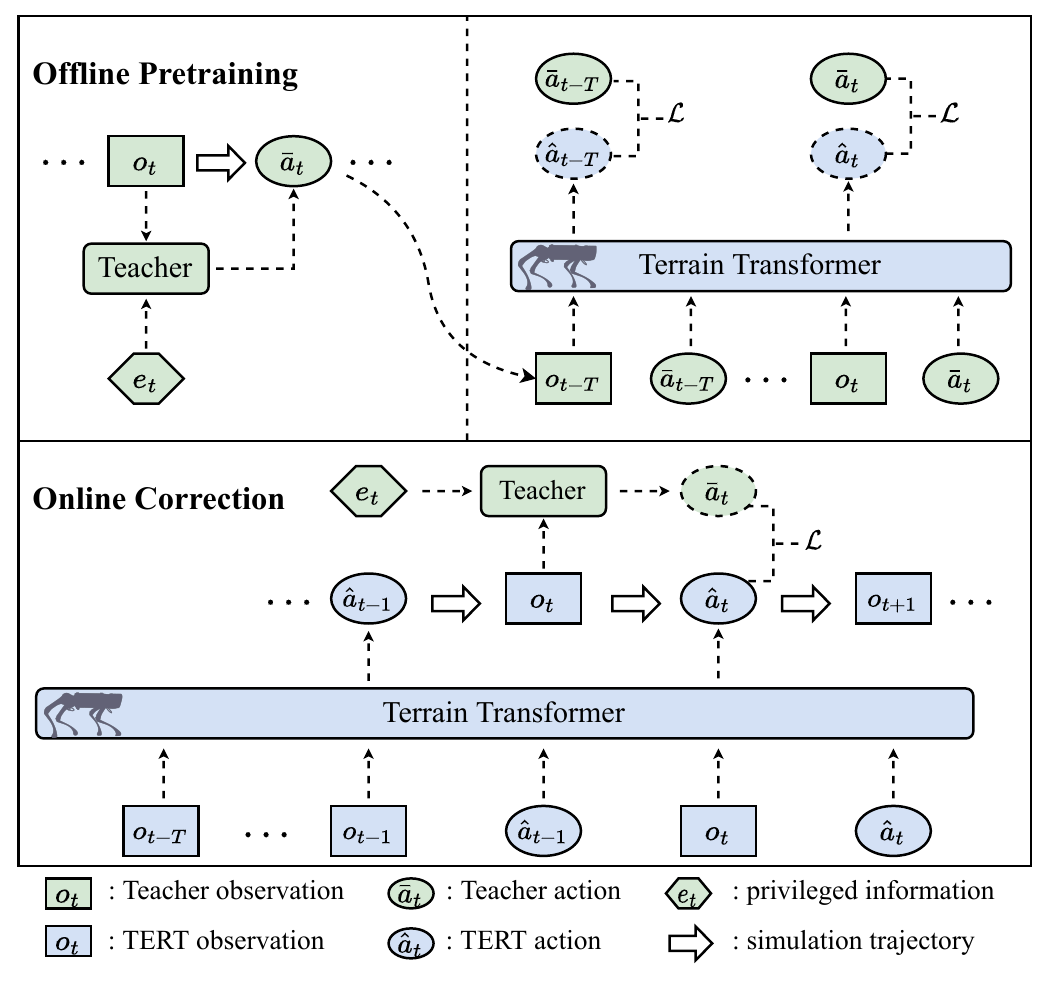}
\vspace{-25pt}	
 \caption{Illustration of Terrain Transformer training framework. Dashed lines represent actions not executed in the simulator.}
	\vspace{-15pt}
	\label{fig:framework}
\end{figure}
\begin{equation}
\begin{aligned}
    \mathcal{L} &= \sum_{t=1}^{T} (\hat{a}_t - \bar{a}_t)^2,
    \\     (\hat{a}_1, \hat{a}_2, \dots, \hat{a}_T) &\sim \mathrm{Trans}(o_1,\bar{a}_1,\dots, o_T,\bar{a}_T),
\end{aligned}
\end{equation}
where $(o_t, \bar{a}_t)$ is the observation-action pair in the teacher's trajectory.

\vspace{0.1cm}
\noindent \textbf{Online Correction.} However, a Transformer controller that achieves small prediction losses on teacher's trajectories does not perform well when tested in the simulator since the input distribution, \ie the observation-action sequences it encounters, differs from that in the training dataset. 
Inspired by the DAgger (Data Aggregator) algorithm \cite{dagger}, which queries an expert for action labels of on-policy data, we propose the 
following online correction stage to address the input distribution shift issue. Specifically, in the online correction stage, the Transformer will then be used to interact with the simulator while the teacher policy gives the target action at the same time. Afterward, the Transformer is trained to predict the teacher's action $\bar{a}_t$ conditioned on its own trajectories:
\begin{equation}
\begin{aligned}
    \mathcal{L} &= \sum_{t=1}^{T} (\hat{a}_t - \bar{a}_t)^2,
    \\         (\hat{a}_1, \hat{a}_2, \dots, \hat{a}_T) &\sim \mathrm{Trans}(o_1,\hat{a}_1,\dots, o_T,\hat{a}_T).
\end{aligned}
\end{equation}

Both these two stages play an essential role in the performance of TERT. Without the offline pretraining stage, the initial Transformer controller can hardly give reasonable control commands, and the on-policy data will be restricted and low-quality, thus affecting the efficiency and effectiveness of the algorithm. On the other hand, without the online correction stage, the Transformer model will suffer from the input distribution shift problem, as mentioned before. More experimental comparison of these two variants is provided in Section~\ref{sec:Empirical Study}.


%% file: evaluation.tex
\section{Experimental Results}
\label{sec:Experimental Result}
Our experiments aim to answer the following questions:
\begin{itemize}
	\item How does TERT perform compared with previous state-of-the-art methods in multi-terrain quadrupedal locomotion?
        \item What are the most valuable components of our overall algorithm?
	\item Does TERT benefit from the capability and self-attention mechanism of Transformer to model long sequences?
	\item Could the achievement of TERT in simulation be well transferred to real robots?
\end{itemize}
To answer these questions, we apply our proposed approach to the Unitree A1 robot \cite{unitree}, which weighs around 12 kg and has 12 motors in total (two for each hip joint and one for each knee joint).

\subsection{Experimental Setup}
\label{sec:Experimental Setup}
We implement TERT and baselines based on the open-source codebase in Rudin \etal~\cite{leggedgym}, which leverages the Isaac Gym simulator \cite{isaacgym} to support simulation of thousands of robots in parallel and provides multi-terrain simulation, including slopes, stairs, and discrete obstacles. Here we briefly describe some settings that are important for our experiments and defer other details to the original paper.
\vspace{0.1cm}
\newline
\noindent \textbf{Observation and Action.} Precisely, the proprioception observation $o_t \in \mathbb{R}^{48}$ consists of: base linear and angular velocities, gravity projection, commands, linear and angular velocities of joints, and last action $a_{t-1}$. The action $a_t \in \mathbb{R}^{12}$ specifies the target positions for each joint, which are then converted to joint torques through a PD controller:
\begin{equation}
    \mathrm{torque}=k_{p}\left(q_{d}-q\right)+k_{d}\left(\dot{q}_{d}-\dot{q}\right),
\end{equation}
where $q$ and $q_d$ are the current position and target position for the joint, respectively, and $(k_p,k_d)$ are the hyper-parameters for the PD controller. 

\vspace{0.1cm}
\noindent \textbf{Reward Function.} We adopt a reward function similar to Rudin \etal~\cite{leggedgym}, which encourages the robot to follow commanded velocities and penalizes velocities along other axes, large joints torques, accelerations and collisions. We additionally add two reward terms to penalize large action magnitude and encourage smoothness of joint torques \cite{rma}, which we find helpful for sim-to-real transfer.

\vspace{0.1cm}
    \noindent \textbf{Model Architecture.} For teacher policy, the privileged information $e_t$ is first encoded by a three-layer MLP with 256 hidden units into a latent vector $l_t \in \mathbb{R}^{12}$. Then $l_t$ is concatenated with $o_t$ before being passed to a three-layer MLP with (512, 256, 128) hidden units. For Transformer, we use 3 self-attention layers with embedding size 256. The dropout rate is set to 0.05 and the trajectory length $T$ is set to 20.



\begin{table*}[t]
\vspace{5pt}
\setlength{\tabcolsep}{5.8pt}
\caption{
Comparison results on different terrains in terms of return, smoothness, and energy consumption. Results are averaged over 1000 trajectories with different difficulties. Bold numbers indicate the best scores among algorithms excluding teacher.}
\centering
\small
\begin{tabular}{ccccccccc}
\toprule
\multicolumn{1}{c}{} & \multicolumn{1}{c}{\bf Terrain} & \multicolumn{1}{c}{\bf Teacher} & \multicolumn{1}{c}{\bf TERT (ours)} & \multicolumn{1}{c}{\bf RMA} & \multicolumn{1}{c}{\bf PPO} & \multicolumn{1}{c}{\bf StackedPPO} & \multicolumn{1}{c}{\bf GRU} \\
\midrule
\multirow{6}{*}{Return} & Smooth Slope &  $21.33 \pm 2.40$ &  $\bf{21.26} \pm 2.51$ &  $21.11 \pm 2.71$ &  $17.30 \pm 2.67$ & $19.97 \pm 2.69$ & $20.30 \pm 2.74$ \\
 & Rough Slope     &  $20.38 \pm 2.64$ &  $20.11 \pm 3.15$ &  $\bf{20.21} \pm 2.74$ &  $16.27 \pm 2.89$ & $19.28 \pm 2.90$ & $19.54 \pm 2.73$ \\
 & Stair Up  &  $20.49 \pm 1.24$ &  $\bf{19.66} \pm 2.26$ &  $19.20 \pm 2.64$ &  $16.43 \pm 2.38$ & $18.82 \pm 2.13$ & $17.69 \pm 3.02$ \\
 & Stair Down  &  $20.15 \pm 3.67$ &  $\bf{19.26} \pm 4.49$ &  $18.79 \pm 4.78$ &  $17.36 \pm 4.26$ & $19.08 \pm 4.10$ & $19.13 \pm 4.04$ \\
 & Obstacle &  $22.10 \pm 1.98$ &  $\bf{21.97} \pm 2.28$ &  $21.71 \pm 2.35$ &  $17.34 \pm 2.72$ & $20.61 \pm 1.95$ & $20.55 \pm 2.62$ \\
 
  & Average &  $20.89$ &  $\bf{20.43}$ &  $20.20 $ &  $16.94$ & $19.55$ & $19.44$ \\
\midrule
\multirow{6}{*}{Smooth} & Smooth Slope &  $1.30 \pm 0.14$ &  $\bf{1.26} \pm 0.12$ &  $1.75 \pm 0.15$ &  $1.71 \pm 0.52$ & $1.55 \pm 0.69$ & $1.38 \pm 0.25$ \\
 & Rough Slope  &  $1.35 \pm 0.16$ &  $\bf{1.31} \pm 0.14$ &  $1.82 \pm 0.18$ &  $1.77 \pm 0.64$ & $1.67 \pm 1.03$ & $1.48 \pm 0.19$ \\
 & Stair Up  &  $1.36 \pm 0.08$ &  $1.38 \pm 0.11$ &  $1.80 \pm 0.16$ &  $2.09 \pm 0.49$ & $1.70 \pm 0.54$ & $\bf{1.32} \pm 0.28$ \\
 & Stair Down &  $1.37 \pm 0.22$ &  $\bf{1.30} \pm 0.22$ &  $1.85 \pm 0.59$ &  $1.68 \pm 0.48$ & $2.23 \pm 1.92$ & $1.57 \pm 0.29$ \\
 & Obstacle &  $1.25 \pm 0.14$ &  $\bf{1.21} \pm 0.10$ &  $1.71 \pm 0.20$ &  $1.72 \pm 0.65$ & $1.44 \pm 0.51$ & $1.30 \pm 0.20$ \\
 & Average &  $1.32$ &  $\bf{1.29}$ &  $1.78$ &  $1.79$ & $1.72$ & $1.41$ \\
\midrule
\multirow{6}{*}{Energy} & Smooth Slope &  $39.20 \pm 5.29$ &  $38.93 \pm 5.75$ &  $\bf{38.91} \pm 5.69$ &  $55.40 \pm 10.85$ & $41.67 \pm 6.96$ & $42.27 \pm 6.29$ \\
 & Rough Slope  &  $39.57 \pm 6.07$ &  $\bf{38.86} \pm 6.78$ &  $39.97 \pm 6.68$ &  $56.41 \pm 13.01$ & $43.67 \pm 7.51$ & $44.34 \pm 5.87$ \\
 & Stair Up &  $44.43 \pm 4.06$ &  $44.57 \pm 7.65$ &  $41.62 \pm 10.54$ &  $65.74 \pm 14.45$ & $47.69 \pm 7.64$ & $\bf{40.11} \pm 12.01$ \\
 & Stair Down &  $37.62 \pm 8.03$ &  $\bf{37.09} \pm 10.61$ &  $40.37 \pm 11.39$ &  $55.01 \pm 12.98$ & $47.23 \pm 14.20$ & $48.21 \pm 9.59$ \\
 & Obstacle  &  $39.43 \pm 5.23$ &  $\bf{38.61} \pm 5.59$ &  $39.06 \pm 5.94$ &  $56.33 \pm 12.30$ & $40.47 \pm 6.59$ & $39.74 \pm 5.20$ \\
 & Average &  $40.05$ &  $\bf{39.61}$ &  $39.98$ &  $57.77$ & $44.15$ & $42.93$ \\
\bottomrule
\end{tabular}
\label{table:overall comparison}
\end{table*}


\begin{figure*}[htp]
\centering
        \vspace{-10pt}
    \begin{minipage}[t]{0.355\textwidth}
        \centering
        \includegraphics[width=1\textwidth]{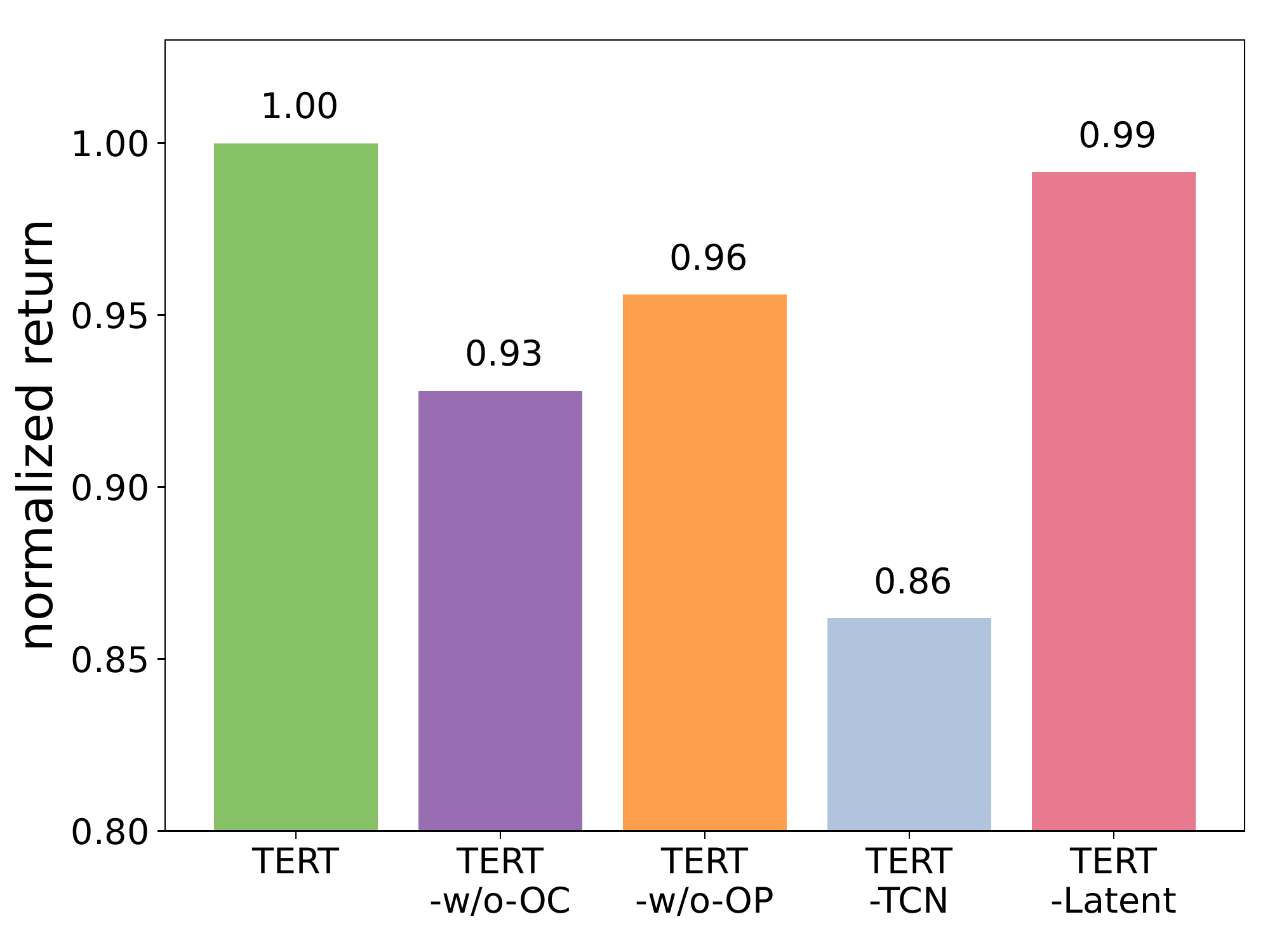}
        \vspace{-22pt}
        \caption{Average normalized return between [0, 1] of TERT and its ablated variants.}
        \label{fig:ablation study}
        \end{minipage}
            \hspace{4pt}        
    \begin{minipage}[t]{0.62\textwidth}
        \includegraphics[width=1\textwidth]{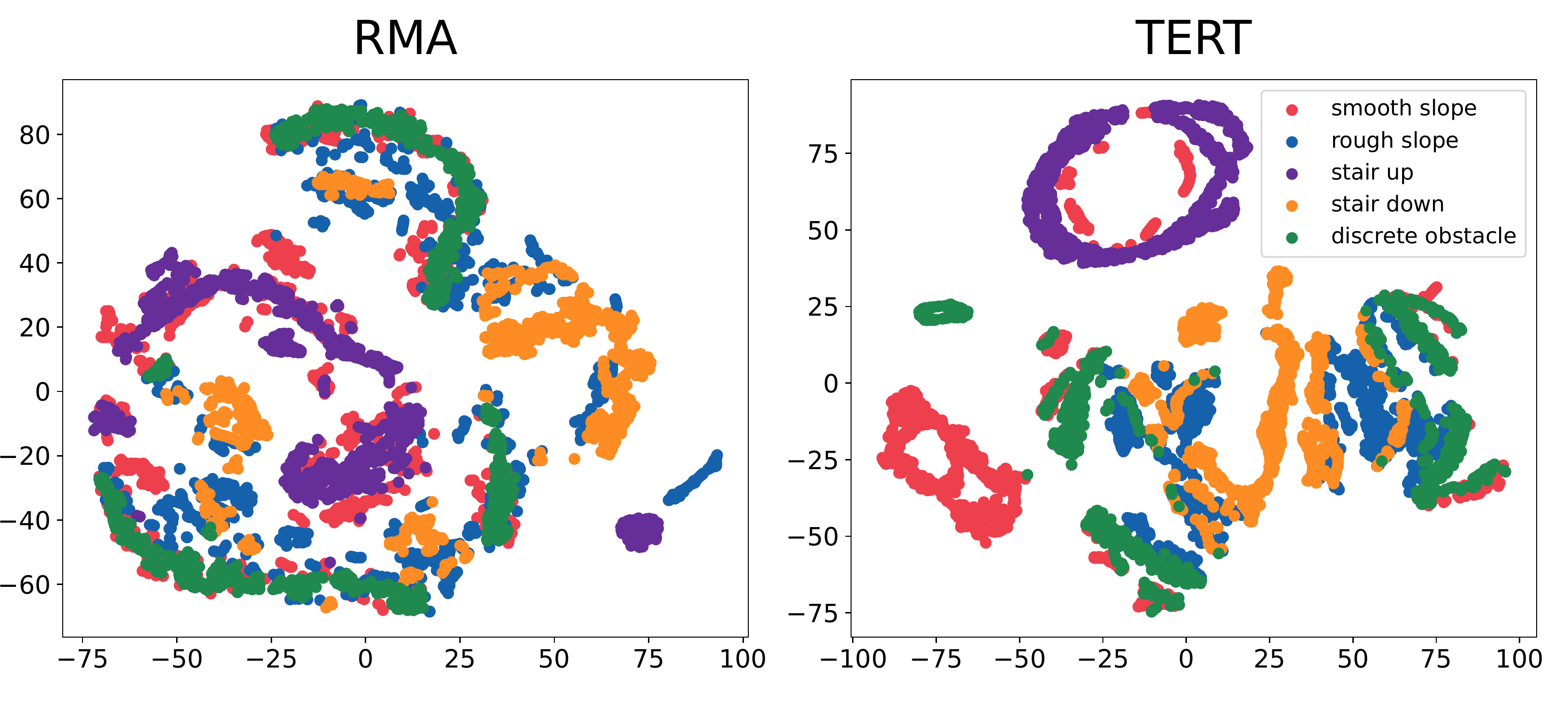}
        \vspace{-22pt}
        \caption{T-SNE visualization for the last hidden layer of RMA and TERT on different terrains. Points of RMA on five terrains largely overlap while TERT holds more clear boundaries.}
        \label{fig:tsne}
    \end{minipage}

        \vspace{-15pt}
\end{figure*}

\subsection{Simulation Comparison}
\label{sec:Simulation Comparison}
We first evaluate our method and baselines in terms of return, control smoothness and energy consumption over different terrains in simulation, where the control smoothness is defined as the average difference between $a_t$ and $a_{t-1}$. The baselines we compared include:
\begin{itemize}
	\item \textbf{Teacher.} The teacher policy with privileged information serves as an oracle.
	\item \textbf{RMA.} We implement the student policy according to the RMA algorithm \cite{rma}, which trains a Temporal Convolutional Network (TCN) to estimate $l_t$ using the last 50 steps (on-policy data) and removes the reliance on predefined trajectory generator used in Lee \etal~\cite{terrain}. 
	\item \textbf{PPO.} A policy directly trained via PPO with $o_t$ as input.
	\item \textbf{StackedPPO.} PPO policy with stacked historical observations and actions as input. Note that PPO can be viewed as a special case of StackedPPO with the stacked length set to 1.
	\item \textbf{GRU.} Replacing the multi-layer perceptron (MLP) in PPO with the GRU \cite{GRU} model.
\end{itemize}
\noindent To ensure a fair comparison, we use the same teacher policy, \ie the one listed in Table~\ref{table:overall comparison}, to train both RMA and TERT.

The comparison results are shown in Table~\ref{table:overall comparison}. Our method TERT outperforms other baselines in terms of return on most terrains except rough slope, especially in more challenging scenarios such as stair up and stair down. The gap between TERT and teacher is smaller than that between RMA and teacher, verifying the advantage of Transformer in sequence modeling compared to traditional models such as TCN. Moreover, Stacking past observations and actions or processing them with GRU helps address the partially observable problem but is not comparable to TERT or RMA without the guidance of privileged information. In addition, TERT achieves lower energy consumption and smoother control strategy, though our reward function does not penalize energy consumption explicitly.

\subsection{Empirical Study}
\label{sec:Empirical Study}
In this section, we provide empirical studies to understand the importance of each component and the benefit our method derives from the Transformer architecture.

\vspace{0.1cm}
\noindent \textbf{Ablation Study.} We first conduct experiments to evaluate the main components of our algorithm, \ie the two-stage training framework, the Transformer model, and the end-to-end imitation scheme. Specifically, we compare the original TERT with four variants: i) \textbf{TERT-w/o-OC.} Variant of TERT without online correction (OC) training stage. ii) \textbf{TERT-w/o-OP.} Variant of TERT without offline pretraining (OP) training stage. iii) \textbf{TERT-TCN.} Replacing Transformer in TERT with TCN model. iv) \textbf{TERT-Latent.} Utilizing Transformer to estimate the latent vector like RMA instead of imitating the teacher's action end-to-end. Results are shown in Figure~\ref{fig:ablation study}. We find that ablating either the offline pretraining or online correction stage decreases the performance to some extent, validating the effectiveness of our proposed training framework, as discussed in Section~\ref{sec:Transformer Learning}. Furthermore, removing the Transformer causes severe performance degradation, while using Transformer to estimate the latent vector retains most of the advantages of TERT, which further reveals the superiority of Transformer architecture in sequence modeling.

\vspace{0.1cm}
\noindent \textbf{Hidden Layer Visualization.} We then unroll TERT and the RMA baseline over different terrains and visualize their last hidden layer in Figure~\ref{fig:tsne}. As the t-SNE result shows, the points of RMA on five terrains largely overlap, while our method holds more clear boundaries. We can also observe a slight overlap for TERT on rough slope and discrete obstacle since both terrains need to tackle uneven ground. This result helps explain why RMA performs excellently on some terrains, such as smooth and rough slopes, but not so well on others. In contrast, TERT can leverage the high-capacity Transformer to better capture the characteristics of the environment and model diverse behaviors, which may help generalize to different terrains.


\begin{figure*}[tb]
	\centering
	\vspace{5pt}
\includegraphics[width=1.0\textwidth]{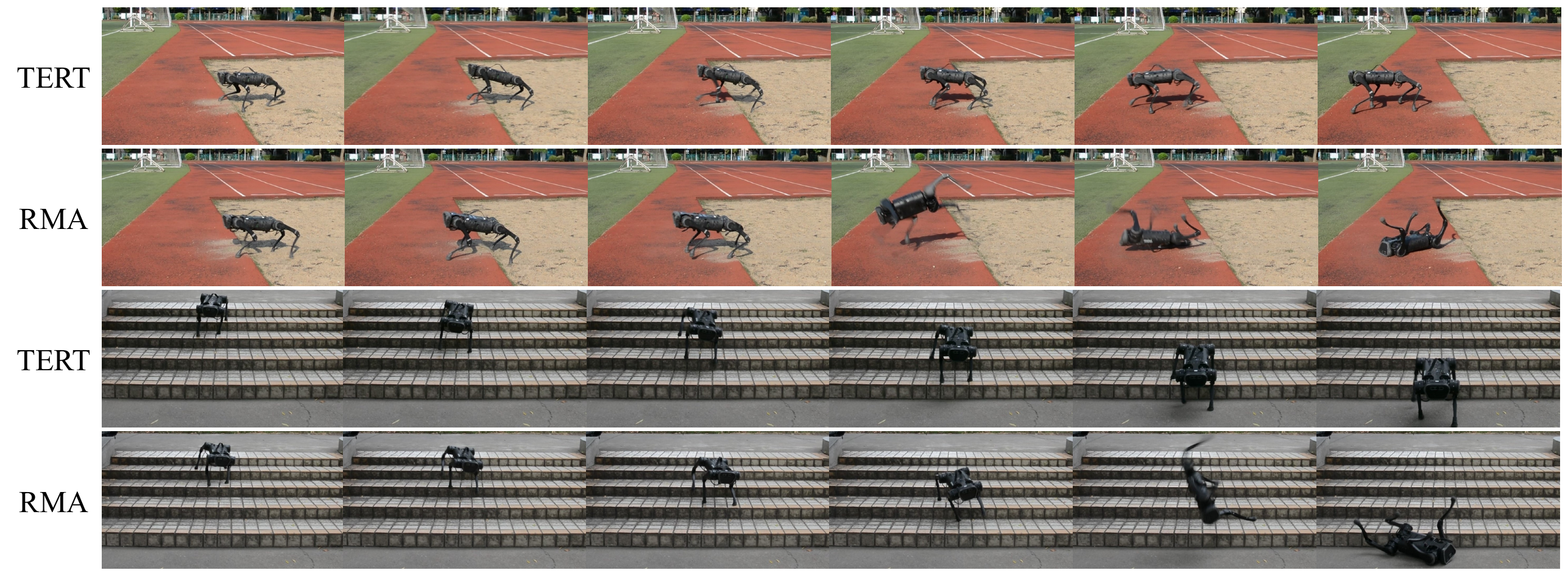}
	\vspace{-23pt}
	\caption{Snapshots of TERT and RMA traversing across sand pit (top) and stair down (bottom) terrain.}
	
	\vspace{-10pt}
	\label{fig:snapshots}
\end{figure*}

\begin{figure}[htb]
	\centering
\vspace{-5pt}
\includegraphics[width=0.5\textwidth]{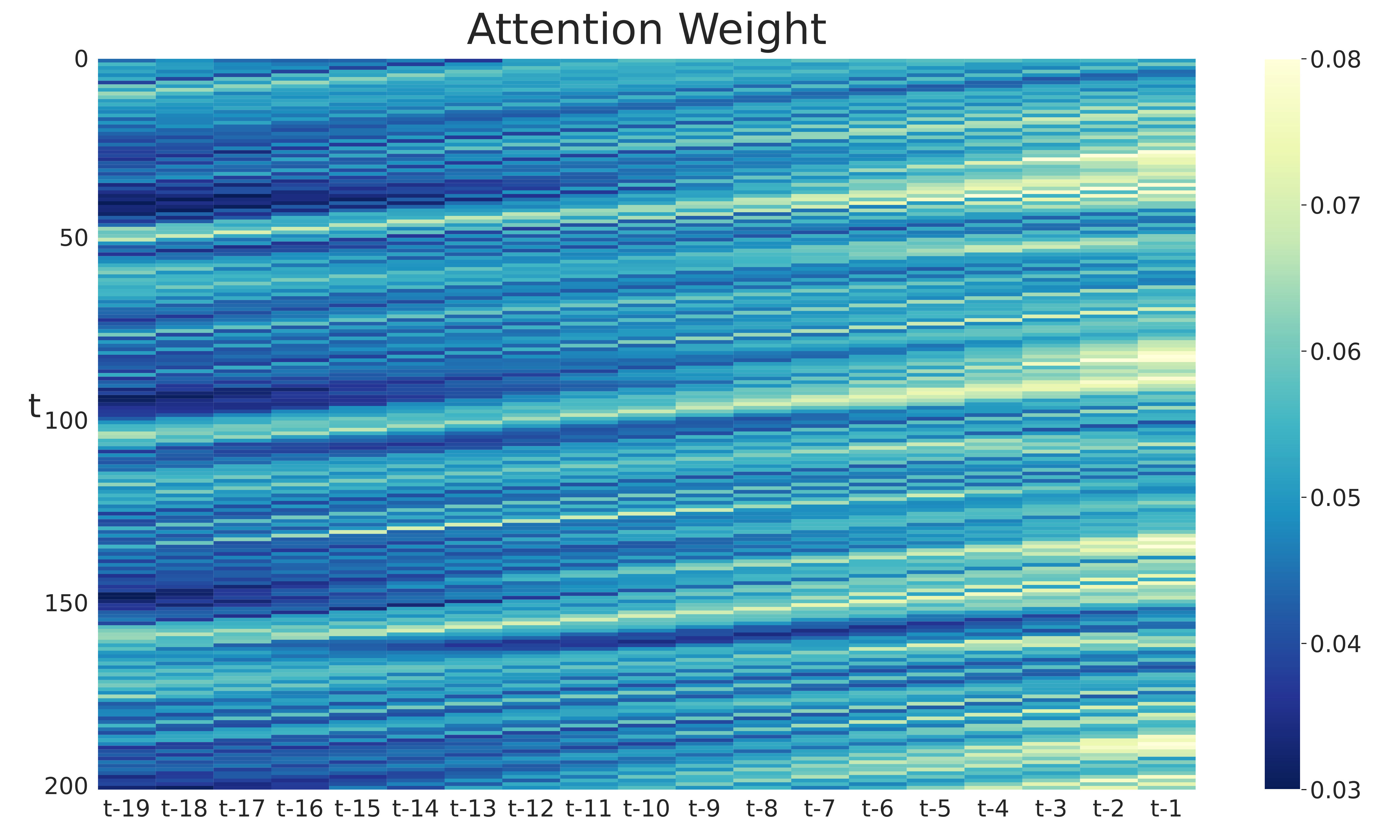}
	\vspace{-20pt}
	\caption{Heatmap of Transformer attention weight on stair down terrain. Y-axis indicates different timestep $t$, and x-axis indicates the corresponding history sequences (20 timesteps in our setting). For example, the color of the square at $50$-th row and $10$-th column represents the attention weight Transformer assigns for $(o_{40},a_{40})$ to predict $a_{50}$.}
	
	\vspace{-15pt}
	\label{fig:attention}
\end{figure}

\vspace{0.1cm}
\noindent \textbf{Attention Weight.} The self-attention mechanism endows Transformer models with the ability of context-dependent control \cite{TrMRL}, which is important for multi-terrain locomotion. We plot the attention weights along a stair-down trajectory of 200 timesteps in Figure~\ref{fig:attention}. As the figure shows, TERT allocates different attention weights over time. Notice that weights assigned to earlier timesteps (\eg $t-19$) are comparable to those assigned to later timesteps, indicating that Transformer can make good use of long historical information. Besides, the attention weight shows a cyclical pattern, reflecting the corresponding periodic gait of the quadrupedal robot.


\subsection{Real-world Application}
\label{sec:Real-world Application}
We apply TERT and RMA policy to an A1 robot without any fine-tuning in the real world. Both methods are directly run on the A1 hardware without external computing devices. The base linear velocities in $o_t$ are estimated from the accelerometer, and other proprioception information is read from the IMU and other sensors. We set the control frequency to 50Hz and $k_p = 55, k_d = 8$ for both methods.

\begin{table}[h]
\caption{Real-world traversing success rates on different terrains. Results are averaged over five trials to reduce damage to the robot hardware.}
\centering
\small
\begin{tabular}{cccc}
\toprule
 & Sand Pit & Stair Down  & Other Terrains  \\
\midrule
    \multicolumn{1}{c}{\bf TERT (ours)}  &  $\bf{100\%}$ & $\bf{60\%}$  &  ${100\%}$  \\
     \multicolumn{1}{c}{\bf RMA}  & $0\%$ &  $0\%$ &  ${100\%}$ \\
\bottomrule
\end{tabular}
\label{table:real world success}
\end{table}

We select nine different terrains for a comprehensive evaluation, including grass, soil, pebble, smooth floor, damp wood, slope up, slope down, sand pit and stair down \footnote{We do not include stair up terrain in the comparison since it is too difficult for both methods to accomplish it in the real world without elevation information \cite{rma}.}. The success rates are listed in Table~\ref{table:real world success}. Both methods can pass through the first seven terrains with a 100$\%$ success rate, while RMA fails on sand pit and stair down without exception. We attribute this failure to the sensitivity of RMA to the estimated latent vector.
In contrast, our method exhibits a more stable control behavior and successfully traverses more challenging terrains, only fails sometimes on stair down. We provide snapshots of trials on sand pit and stair down in Figure~\ref{fig:snapshots}. Please refer to the supplemental video for detailed comparisons on more terrains.

%% file: conclusion.tex
\section{Conclusion}	
\label{sec:Conclusion}
In this paper, we propose Terrain Transformer (TERT), a simple yet effective method to leverage Transformer for quadrupedal locomotion over multiple terrains, including a two-stage training framework to incorporate Transformer with privileged learning. Extensive simulation and real-world experiments demonstrate that TERT outperforms the state-of-the-art baseline on different challenging terrains. Further empirical analyses provide insight that the main superiority of TERT lies in the high capacity for sequence modeling and the self-attention mechanism of Transformer. We believe our work takes an important step towards application of Transformer in quadruped robot locomotion, which could become a new paradigm of sim-to-real transfer for robot control. For future work, it is appealing to integrate elevation information from LiDAR or camera into our framework and investigate its extension to multi-agent control tasks. 

%% file: ack.tex
\section*{Acknowledgements}	
\label{sec:Acknowledgements}
The SJTU Team is supported by Shanghai Municipal Science and Technology Major Project (2021SHZDZX0102) and National Natural Science Foundation of China (62076161).
The author Hang Lai is supported by Wu Wen Jun Honorary Doctoral Scholarship, AI Institute, Shanghai Jiao Tong University.

%% file: appendix.tex
\newpage
\appendix

\subsection{Training Details}
\begin{table}[th]
\caption{Domain Randomization Range}
\vspace{-10pt}
\label{tabel:dr range}
\begin{center}
\begin{tabular}{lcc}
\toprule
 \textbf{Parameter} & \textbf{Training Range} & \textbf{Testing Range} \\
 \midrule
 Friction & [0.5, 1.25]  & [0.1, 2.0]  \\
 Added mass (Kg)  & [0, 5]  & [0, 7]  \\
 $k_p$ & [45, 65] & [40, 70]  \\
 $k_d$  &  [0.7, 0.9] & [0.6, 1.0] \\
\bottomrule
\end{tabular}
\vspace{-10pt}
\end{center}
\end{table}

\vspace{0.1cm}
    Following Rudin \etal~\cite{leggedgym}, we create 4096 environment instances to collect data in parallel. Each environment is chosen from five types of terrain (smooth slope, rough slope, stair up, stair down, discrete obstacle) with different difficulty levels. In each environment, the robot is initialized with random poses and commanded to walk forward at a speed of 0.4m/s. The robot receives new observations and updates its actions every 0.02 seconds. An episode is terminated if the robot trunk touches the ground or 1000 timesteps (20 seconds in simulation) are collected. We also randomize the physical parameters to further improve the robustness of the policy. The domain randomization range is listed in Table~\ref{tabel:dr range}. We train the teacher policy for 20k iterations; each iteration corresponds to 98304 (4096$\times$24) timesteps. At each stage of Transformer training, we collect a dataset of 20 million timesteps and then train the Transformer for 200k updates with batch size 64.

\subsection{Sequence Length}

\begin{figure}[htb]
	\centering
\vspace{-5pt}
\includegraphics[width=0.50\textwidth]{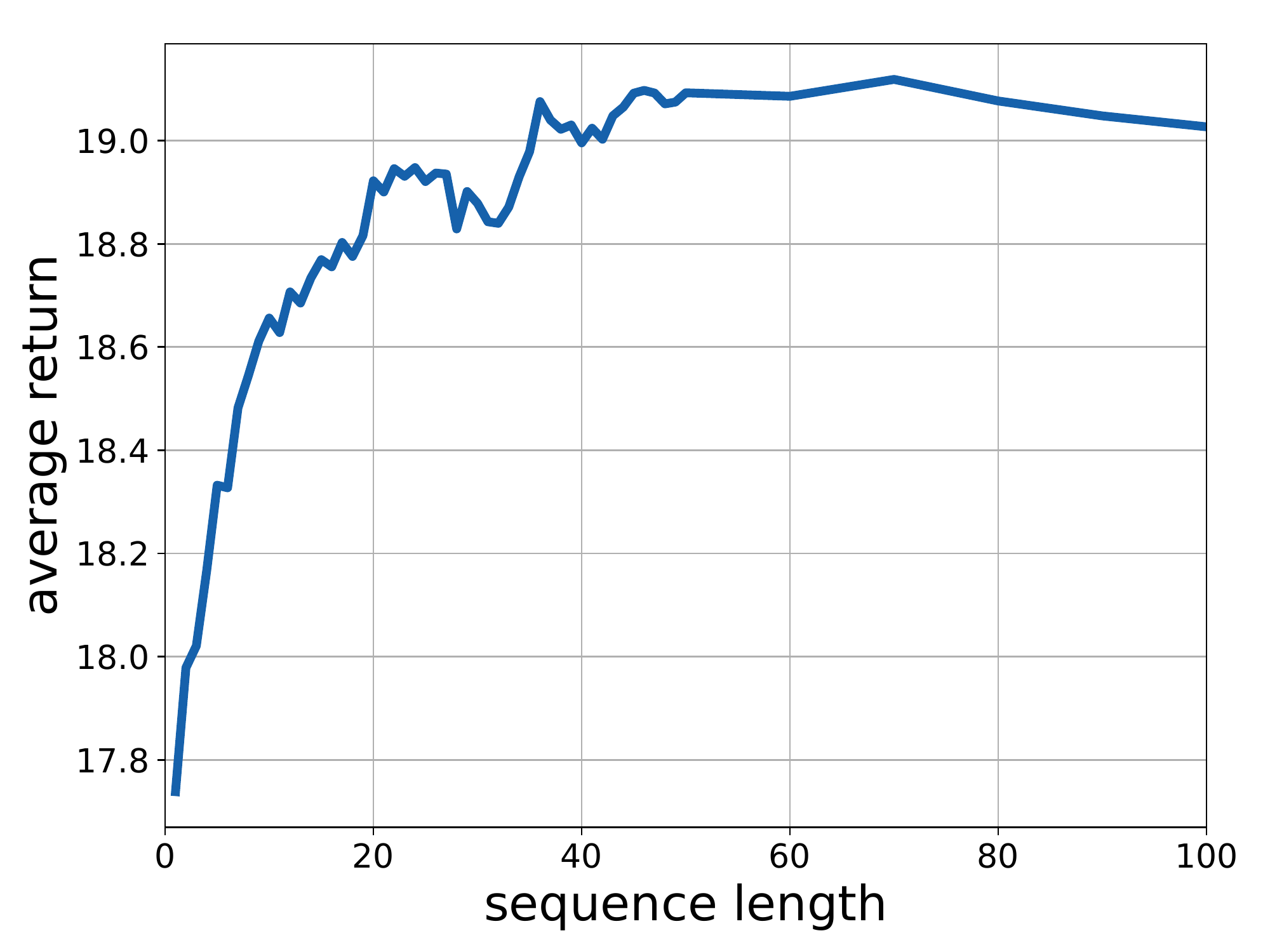}
	\vspace{-10pt}
	        \caption{Average return of TERT after offline pretraining with different sequence length $T$.}
        \label{fig:sequence length}
	\vspace{-5pt}
\end{figure}

\vspace{0.1cm}
\noindent \textbf{Sequence Length.} We further conduct experiments of TERT with different sequence length $T$.
Results are shown in Figure~\ref{fig:sequence length}. Note that when $T = 1$, TERT degenerates into one-step behavior cloning. We observe that up to a certain level, increasing the sequence length $T$ yields better performance, which reveals the advantage of longer sequence modeling. However, too large $T$ will affect the computation efficiency of the algorithm, thereby decreasing the control frequency of the robot. 
Therefore to trade off these two terms, we choose $T=20$ in our experiments, which achieves acceptable performance according to Figure~\ref{fig:sequence length}.